\documentclass{article} 
\PassOptionsToPackage{table}{xcolor}
\usepackage{iclr2027_conference,times}
\iclrfinalcopy 

\usepackage{amsmath,amsfonts,bm}

\def\eqref#1{equation~\ref{#1}}

\def\1{\bm{1}}

\DeclareMathAlphabet{\mathsfit}{\encodingdefault}{\sfdefault}{m}{sl}
\SetMathAlphabet{\mathsfit}{bold}{\encodingdefault}{\sfdefault}{bx}{n}

\usepackage{hyperref}
\usepackage{url}

\usepackage{graphicx} 
\usepackage{amsmath}
\usepackage{amssymb}
\usepackage{booktabs}
\usepackage{multirow}
\usepackage{array}
\usepackage{tabularx}
\usepackage[table]{xcolor}
\usepackage{float}
\usepackage{titlesec}   
\usepackage{tcolorbox}  
\tcbuselibrary{skins,breakable}
\usepackage{enumitem}   
\usepackage[font={small,it}]{caption} 

\definecolor{cotgenrow}{HTML}{EAF2FC}
\definecolor{cotcondrow}{HTML}{FCEAEA}
\definecolor{deltapos}{HTML}{1B8A3A}
\definecolor{deltaneg}{HTML}{C0392B}
\newcommand{\dpos}[1]{\textcolor{deltapos}{#1}}
\newcommand{\dneg}[1]{\textcolor{deltaneg}{#1}}

\newtcolorbox{promptbox}[1][]{
  enhanced, breakable, colback=black!4, colframe=black!40,
  boxrule=0.4pt, left=6pt, right=6pt, top=4pt, bottom=4pt,
  fontupper=\small\ttfamily,
  #1
}

\newcommand{\methodname}{\textsc{Eklav}}
\newcommand{\Methodname}{\textsc{Eklav}}
\newcommand{\baselinename}{\textsc{std-CoT}}
\newcommand{\Baselinename}{\textsc{std-CoT}}
\newcommand{\naivename}{\textsc{Naive-SFT}}
\newcommand{\Naivename}{\textsc{Naive-SFT}}
\newcommand{\maskonlyname}{\textsc{Mask-only}}
\newcommand{\Maskonlyname}{\textsc{Mask-only}}

\title{EKLAV: Elicited Key CoT Learning via Answer only superVision}

\author{\mdseries
  $^{*}$Adarsh Singh$^{1}$ \quad $^{*}$Aman Singh$^{1}$ \quad $^{*}$Kushal Raj Bhandari$^{2}$ \\
  Soham Dan$^{3}$ \quad Anoop Kumar$^{4}$ \quad Vivek Gupta$^{1}$ \\[4pt]
  {\small $^{1}$Arizona State University \quad $^{2}$Rensselaer Polytechnic Institute \quad $^{3}$Scale AI \quad $^{4}$Capital One} \\
  {\small \texttt{\{asing725,asing651\}@asu.edu} \quad \texttt{bhandk@rpi.edu} \quad \texttt{anoop.kumar@capitalone.com}} \\
  {\footnotesize $^{*}$Equal contribution}
}

\begin{document}

\maketitle
\lhead{Preprint} 

\begin{abstract}
  Reasoning distillation trains a student to reproduce both the teacher's chain-of-thought trace and its final answer.
  We ask whether the trace is better used as privileged training context than as an imitation target.
  \Methodname{} keeps the teacher's reasoning in the input sequence but supervises only the final task answer, applying a task-dependent placement that moves part or all of the trace into the prompt.
  On BRIGHT passage reranking, \methodname{} improves average nDCG@10 at all four tested checkpoints across two model families, winning 41 of 48 model-domain comparisons while reducing training compute by up to 31.6\% and inference time by up to 32.7\%.
  On four held-out table-reranking benchmarks, it is the best of four conditions on three, exceeding full-trace distillation by 3.0 points and answer-only training without reasoning by 5.5 points on average.
  Analysis of generated traces reveals that \methodname{} students defer their first verbalized verdict later (0.76 vs.\ 0.44 relative position), reverse fewer decisions (5\% vs.\ 16\%), and produce reasoning more predictive of their own rankings than \baselinename{}'s, though less predictive than the base model's.
  The recipe reuses a fixed teacher corpus, requires no student rollouts or iterative teacher feedback, and applies to both pointwise and listwise formulations.
  Code and Data are at \url{https://github.com/AdarshSingh7647/Eklavya}.
\end{abstract}

\section{Introduction}
\label{sec:intro}
\vspace{-0.5em}
Reasoning distillation transfers the capabilities of a large teacher model to a smaller student through supervised fine-tuning \cite{guoDeepSeekR1IncentivizesReasoning2025,muennighoffS1SimpleTesttime2025,yeLIMOLessMore2025}. The standard recipe, which we call \baselinename{}, trains the student to reproduce both the teacher's reasoning trace and its final answer, implicitly treating the intermediate reasoning as a desirable supervision target even though fluent chain-of-thought need not faithfully describe the computation behind a prediction \cite{wangSCOTTSelfconsistentChainofthought2023,arcuschinChainofThoughtReasoningWild2026}. Separating rationale supervision from answer supervision has been shown to improve out-of-domain generalization \cite{daiImproveStudentsReasoning2024}, and more broadly, chain-of-thought reasoning has been characterized as a brittle artifact of the training distribution: controlled studies report near-perfect in-distribution performance that collapses under moderate task, length, or format shifts \cite{zhao-etal-2026-chain}, and fine-tuning on reasoning traces has been shown to reduce both accuracy and CoT faithfulness outside the fine-tuning task category, more so for smaller models \cite{lobo-etal-2025-impact}. We observe the same asymmetry in our own table-reranking results: \naivename{}, an answer-only baseline with no teacher reasoning at all, outperforms full-trace \baselinename{} on three of four out-of-domain benchmarks we test, even though \baselinename{} is the stronger of the two on every in-domain benchmark (Section~\ref{sec:results-table}). These observations raise a direct question: must a student reproduce the teacher's reasoning to benefit from it?

Reranking offers a controlled setting for studying this question. A reranker must reason about the relevance of a candidate document to a query, while ultimately producing a well-defined and verifiable ranking decision. This structure allows us to separate supervision of intermediate reasoning from supervision of the final task output with fewer confounds from stylistic variation or unconstrained generation. BRIGHT \cite{suBRIGHTRealisticChallenging2025} further provides a demanding evaluation across heterogeneous reasoning domains, including scientific literature, mathematics, coding, and logical inference. Performance across these domains therefore tests whether a distillation strategy transfers beyond a narrow reasoning pattern.

We study a simple alternative, \methodname{}, that preserves teacher reasoning as training context while removing direct supervision over the reasoning tokens. For pointwise passage reranking, a filtered prefix of the teacher trace is placed in the prompt as a reasoning hint, while the remaining teacher continuation stays in the assistant sequence with its token-level loss masked. For listwise table reranking, the complete teacher trace is provided in the prompt. In both formulations, teacher reasoning conditions the supervised prediction, but the optimization objective applies only to the final task answer. The model therefore learns from teacher reasoning while never being asked to reproduce it. Both formulations operate entirely on a fixed teacher-generated corpus and require neither student rollouts nor iterative teacher feedback. At evaluation time, all teacher-provided reasoning is removed and the student generates its own reasoning before producing the final prediction.

This formulation connects reasoning distillation to learning with privileged information \cite{vapnikNewLearningParadigm2009,lopezpazUnifyingDistillationPrivileged2016}, where additional information is available during training and withheld at inference. Answer-focused supervision with privileged reasoning context also appears in Masked Distillation \cite{kalwarMaskedDistillationInternalizing2026}. \Methodname{} shares the principle of masking direct supervision over teacher reasoning. It develops this idea for reranking with task-dependent context placement, fully offline training, and both pointwise and listwise formulations, and compares against full-trace distillation and answer-only supervision without reasoning context. We evaluate across four model scales and two model families, complementing task-level results with trace-level analysis of how the supervision objective changes the reasoning the student produces.
\vspace{-0.5em}
\paragraph{Contributions.}
\begin{itemize}[leftmargin=1.2em,itemsep=2pt,topsep=2pt]
\item For reranking, we formulate \methodname{} as answer-only distillation with teacher reasoning retained as privileged conditioning context, separating the effects of reasoning supervision, context placement, and the removal of teacher context at evaluation.
\item Across BRIGHT, \methodname{} improves nDCG@10 at all four tested checkpoints spanning two model families and outperforms full-trace distillation in 41 of 48 model-domain comparisons. Across four held-out table benchmarks, it scores highest among four training conditions on three of four, while reducing unparseable outputs from 0--14.5\% to 0--2.5\%.
\item \Methodname{} reduces training compute by 24--32\% and inference time by 15--33\%, yielding an accuracy-efficiency Pareto improvement at every tested model scale.
\item Trace-level analysis shows that \methodname{} students verbalize their verdicts later, reverse fewer intermediate decisions, and produce reasoning whose content is more predictive of their final rankings than \baselinename{}'s, though less predictive than the base model's.
\end{itemize}

\vspace{-1em}
\section{Related Work}
\label{sec:related-work}
\vspace{-0.75em}

\paragraph{Trace supervision.} Reasoning distillation commonly trains students on teacher-generated traces and answers \cite{guoDeepSeekR1IncentivizesReasoning2025,muennighoffS1SimpleTesttime2025,yeLIMOLessMore2025}. CasCoD decomposes rationale and answer learning into cascaded stages while retaining explicit rationale supervision, reporting out-of-domain gains \cite{daiImproveStudentsReasoning2024}. \Methodname{} removes reasoning supervision entirely and retains the trace only as conditioning context.

\vspace{-0.5em}
\paragraph{Masked Distillation.} Masked Distillation \cite{kalwarMaskedDistillationInternalizing2026} is the most closely related work. In its solution-only setting, teacher feedback conditions on the teacher's reasoning while the student predicts solution tokens. \Methodname{} shares the core idea of answer-focused supervision with reasoning as privileged information but differs in mechanism: it places the teacher trace directly in the student's training sequence, uses hard-target cross-entropy on a fixed corpus, and introduces task-dependent context placement. We evaluate on passage and table reranking rather than mathematical problem solving.

\vspace{-0.5em}
\paragraph{Student-adaptive supervision.} A family of methods makes the supervision depend on the particular student being trained: having the student produce incorrect rationales that the teacher then corrects across multiple rounds \cite{wangDemocratizingReasoningAbility2023}, steering trace generation mid-stream by scoring teacher continuations by perplexity \cite{heStudentintheLoopChainofThoughtDistillation2026}, or retrieving skill-conditioned teachers that score the same student rollout \cite{huangSkillConditionedGatedSelfDistillation2026}. These achieve adaptivity at the cost of additional student or teacher evaluations per example, from multiple teacher samples to full student rollouts. \Methodname{} trades this adaptivity for simplicity: it reuses a precomputed corpus with no student rollouts or iterative feedback, making it applicable to any existing reasoning-distillation dataset.

\vspace{-0.5em}
\paragraph{Shorter reasoning.} A separate line of work attacks the verbosity of reasoning traces directly through compression, token skipping, drafting, and budget-aware generation \cite{kangC3oTGeneratingShorter2025,xiaTokenSkipControllableChainofthought2025,xuChainDraftThinking2025,hanTokenbudgetawareLLMReasoning2025}, reducing cost but retaining the fundamental relationship that the student still reproduces a reasoning artifact, only a shorter one. \Methodname{} does not explicitly target length, but its students produce shorter traces as a side effect of removing the imitation objective (Section~\ref{sec:efficiency}).

\vspace{-0.5em}
\paragraph{Privileged information.} Learning Using Privileged Information (LUPI) \cite{vapnikNewLearningParadigm2009} and generalized distillation \cite{lopezpazUnifyingDistillationPrivileged2016} formalize the setting where training-time information is unavailable at test time. Context distillation trains a context-free model to match the output distribution of a separate context-conditioned model \cite{askellGeneralLanguageAssistant2021}, the reverse direction from ours: it distills across two models, whereas \methodname{} trains a single model that sees privileged context at train time and not at test time. \Methodname{} instantiates LUPI for reasoning traces: the teacher's chain of thought serves as privileged context during training and is withdrawn at evaluation through the reasoning context shift (Section~\ref{sec:method}).

Table~\ref{tab:taxonomy} summarizes the central distinction: what the student conditions on versus what its objective supervises.

\begin{table}[t]
\centering
\small
\setlength{\tabcolsep}{4pt}
\renewcommand{\arraystretch}{1.10}
\caption{\textbf{Training formulations.} The Masked Distillation row describes its solution-only setting. Context and targets omit chat-template delimiters.}
\label{tab:taxonomy}
\begin{tabularx}{\linewidth}{@{}>{\raggedright\arraybackslash}p{0.21\linewidth}XX@{}}
\toprule
\textbf{Method} & \textbf{Student context for answer prediction} & \textbf{Supervision} \\
\midrule
\baselinename{} & Task input and teacher trace & Teacher reasoning and answer tokens \\
\naivename{} & Task input & Answer tokens only \\
Masked Distillation \cite{kalwarMaskedDistillationInternalizing2026} & Question; teacher reasoning informs feedback & Solution tokens \\
\methodname{} & Task input and teacher trace, with task-dependent placement & Answer tokens only \\
\bottomrule
\end{tabularx}
\vspace{-1.0em}
\end{table}

\vspace{-0.5em}
\section{Reasoning-Conditioned Answer Supervision}
\label{sec:method}
\vspace{-0.5em}

\vspace{-0.5em}
\subsection{Problem Formulation}
\label{sec:problem-formulation}
\vspace{-0.25em}

We assume access to a reasoning-distillation corpus produced by a large teacher model. Each example is a triple $(x,r,a)$: a task input, a teacher reasoning trace, and a final answer. Standard distillation (\baselinename{}) supervises both $r$ and $a$, training the student to reproduce the teacher's reasoning token by token. \Methodname{} keeps $r$ in the training sequence as conditioning context but applies the loss only to $a$:
\begin{equation}
\boxed{
\text{reasoning as imitation target}
\quad\longrightarrow\quad
\text{reasoning as conditioning context}
}
\label{eq:spine}
\end{equation}
The student sees the teacher's reasoning during training and learns from it through attention and gradient flow, but is never asked to reproduce it. How the reasoning is partitioned between prompt and response depends on the task.

\vspace{-0.25em}
\subsection{Task-Dependent Reasoning Transformation}
\label{sec:reasoning-transformation}
\vspace{-0.25em}

Let $\Phi_\tau(r)=(h_\tau(r),c_\tau(r))$ partition the reasoning into prompt-side context $h_\tau(r)$ and masked assistant-side context $c_\tau(r)$ for task $\tau$.

\paragraph{Pointwise passage reranking.} The teacher ends each query--passage judgment with \texttt{true} or \texttt{false}. Write the trace as sentences $r=(r_1,\ldots,r_m)$ and let $k=\lfloor m/2\rfloor$. We set
\begin{equation}
\Phi_{\mathrm{point}}(r)=\bigl(F(r_{1:k}),\;r_{k+1:m}\bigr),
\end{equation}
where $F$ is a lexical filter that removes verdict-bearing sentences (Appendix~\ref{sec:appendix-hint-details}). The filtered prefix $r^H=F(r_{1:k})$ moves into the prompt as a reasoning hint; if filtering empties it, the hint is left empty rather than falling back to unfiltered text, so no verdict-bearing sentence reaches the prompt. The teacher continuation $r^C=r_{k+1:m}$ remains in the assistant response but is excluded from the loss. Only the final verdict token is supervised. The corpus is the public Rank1 training set \cite{wellerRank1TesttimeCompute2025}. The filter targets explicit verdict phrases in the prompt-side hint; the unfiltered continuation $r^C$ still conditions the loss (see the Deployment paragraph below for what this implies about the training signal).

\paragraph{Listwise table reranking.} The teacher compares the candidate tables and emits a ranking as a JSON list. We retain the complete trace in the prompt,
\begin{equation}
\Phi_{\mathrm{list}}(r)=(r,\varnothing),
\end{equation}
and supervise the ranking JSON (see the Deployment paragraph below for how placing the complete trace in the prompt bears on what this training signal can and cannot show). The corpus contains 15{,}674 DeepSeek-R1 traces. Four evaluation benchmarks contribute no examples to this fine-tuning corpus; four additional benchmarks are evaluated in domain (Section~\ref{sec:results-table} and Appendix~\ref{sec:appendix-indomain}).

\paragraph{Context allocation and the prompt ratio.} Let $\rho_\tau(r)=|h_\tau(r)|/|r|$ denote the fraction of teacher reasoning placed in the prompt. The pointwise split gives $\mathbb{E}[\rho_{\mathrm{point}}]\approx 0.5$ and the listwise variant gives $\rho_{\mathrm{list}}=1$. In both cases the full teacher trace is required during training; $\rho$ governs how it is partitioned between prompt and masked response, not whether it is consumed. The pointwise split is by sentence count; after filtering, the token-level split can differ from the midpoint. Both variants are subject to the shared task-specific sequence cutoff, which may truncate the serialized example.

\subsection{Training Sequences and Loss}
\label{sec:selective-supervision}
\vspace{-0.25em}

We write a chat sequence as $[\text{prompt};\text{assistant response}]$, suppressing template delimiters, and use $\oplus$ for concatenation. The three training conditions are
\begin{align}
y^{\mathrm{STD}}&=[x;\,r\oplus a],\\
y^{\mathrm{EK}}&=[x\oplus h_\tau(r);\,c_\tau(r)\oplus a],\\
y^{\mathrm{Naive}}&=[x;\,a].
\end{align}
\Methodname{} changes both the context layout and the loss mask relative to \baselinename{}; \naivename{} removes teacher reasoning entirely. All three use token-level cross-entropy under teacher forcing:
\begin{equation}
\mathcal{L}_m(\theta)=-\frac{1}{|\mathcal{M}_m|}\sum_{t\in\mathcal{M}_m}\log p_\theta\!\left(y_t^{(m)}\mid y_{<t}^{(m)}\right),
\label{eq:loss}
\end{equation}
where $\mathcal{M}_{\mathrm{STD}}=\operatorname{pos}(r)\cup\operatorname{pos}(a)$ and $\mathcal{M}_{\mathrm{EK}}=\mathcal{M}_{\mathrm{Naive}}=\operatorname{pos}(a)$. Under \methodname{}, prompt tokens and assistant-side reasoning tokens receive no direct loss; any automatically inserted empty reasoning block is also masked (Appendix~\ref{sec:appendix-masking}). The experimental comparison evaluates the complete recipe --- context placement, loss masking, and filtering together --- against full-trace distillation.

\paragraph{Isolating the hint: \maskonlyname{}.} For pointwise passage reranking, we additionally train $y^{\mathrm{Mask}}=[x;\,r\oplus a]$ with $\mathcal{M}_{\mathrm{Mask}}=\operatorname{pos}(a)$: the same masked-loss objective as \methodname{}, but with no prompt-side hint, so the complete teacher trace $r$ sits in the masked assistant-side context instead of being split into $h_\tau(r)$ and $c_\tau(r)$. This ablation isolates whether \methodname{}'s gains come from the hint itself or simply from masking the reasoning loss (Section~\ref{sec:results-passage}).

\paragraph{Gradient flow under teacher forcing.}
\label{sec:masking-mechanics}
Masking a token removes its direct loss but not its influence: later answer tokens still attend to its hidden state, so the answer loss can produce a nonzero gradient at masked reasoning positions, in both \methodname{} and \baselinename{} (Appendix~\ref{sec:appendix-masking-mechanics}).

\paragraph{Deployment.} During training, the answer is conditioned on teacher-derived context $C_T=(h_\tau(r),c_\tau(r))$; for listwise table reranking this is the complete teacher trace, which typically ends by stating the ranking, and for pointwise reranking the unfiltered continuation $c_\tau(r)$ can similarly carry the verdict. Conditioning on $C_T$ can therefore make the supervised answer recoverable from context alone, so training-time performance cannot by itself show that the student is reasoning toward the answer rather than locating it. What the answer-only loss does is shape how that reasoning is produced: gradients push the model toward whatever reasoning, conditioned on $C_T$, leads to the correct answer, not toward reproducing the teacher's tokens. At evaluation, $C_T$ is removed entirely: the student receives $x$ alone and generates its own reasoning $\hat{R}_S$ before answering (the \textbf{reasoning context shift}, $C_T\to\hat{R}_S$). Since $C_T$ is never available at test time, the reported gains reflect only what the student learned to do with its own reasoning (Appendix~\ref{sec:appendix-deployment}).

\vspace{-0.75em}
\section{Experimental Setup}
\label{sec:experimental-setup}
\vspace{-0.75em}

\paragraph{Models.} We evaluate on four checkpoints across two architecture families: Qwen3-4B, Qwen3-8B, and Qwen3-14B \cite{yang2025qwen3} from one family, and GLM-Z1-9B \cite{glm2024chatglm} from another. All four are trained on passage reranking, comparing \baselinename{} with \methodname{} at each scale; the \maskonlyname{} ablation, which keeps \methodname{}'s masked-loss training but removes the prompt-side CoT hint, is additionally reported for Qwen3-8B (Table~\ref{tab:bright-domains}). Table reranking is trained at a single scale (Qwen3-8B) with the full four-way comparison: the untrained base model, \naivename{}, \baselinename{}, and \methodname{}. All fine-tuned models use LoRA \cite{huLoRALowrankAdaptation2022a} (rank 32, alpha 64, all linear layers). The \naivename{} and base-model controls, and the \maskonlyname{} ablation, are not run at the other three passage scales.

\vspace{-0.75em}
\paragraph{Training.} All runs use learning rate $10^{-4}$, cosine schedule with 5\% warmup, effective batch size 128, bf16 precision, and two epochs in LLaMA-Factory \cite{zheng2024llamafactory}. Sequence cutoff is 2{,}500 tokens for passage reranking and 32{,}768 for table reranking. Within each task, \methodname{} and \baselinename{} share all optimization hyperparameters and are compared at the same training step.

\vspace{-0.75em}
\paragraph{Benchmarks and metrics.} Passage evaluation uses the twelve BRIGHT domains \cite{suBRIGHTRealisticChallenging2025}, a challenging realistic benchmark where reasoning-based rerankers show the largest gains over lexical baselines. Table evaluation uses four held-out (OOD) benchmarks (AITQA \cite{katsis2021aitqa}, FeTaQA, OTT-QA, OpenWikiTables), drawn from the IBM Research Table-Text IR Evaluation collection \cite{doshi2026tableir}, and four in-domain benchmarks (SQA \cite{iyyer-etal-2017-search}, TAT-QA \cite{zhu-etal-2021-tat}, TabFact \cite{2019TabFactA}, NQ-Tables; Appendix~\ref{sec:appendix-indomain}), drawn from the same benchmark paper as our multi-table training data even though themselves held out of training (Appendix~\ref{sec:appendix-datasets}). This is a distributional notion of domain, not literal appearance in the fine-tuning corpus, and it says nothing about overlap with the base model's pretraining data. One partial exception is OTT-QA, built on top of the training split of HybridQA \cite{chen-etal-2020-hybridqa}, which is part of our training corpus; we evaluate on OTT-QA's own test split (Section~\ref{sec:results-table}). Table candidate pools are built by lexical and dense retrieval fused via RRF (Appendix~\ref{sec:appendix-datasets}), following Rank1's approach \cite{wellerRank1TesttimeCompute2025}, so that pool construction does not favor either training or held-out sources and reranking is compared fairly across conditions. We report nDCG@10 throughout, scaling passage scores by 100 for readability.

\vspace{-0.75em}
\paragraph{Passage scoring.} The student generates a reasoning trace $\hat r$ for query $q$ and passage $d$, then scores the pair as $s(q,d)=p_\theta(\texttt{true}\mid q,d,\hat r)$, a pointwise formulation inspired by Rank1 \cite{wellerRank1TesttimeCompute2025}. We rerank Rank1's released BM25 \cite{robertson1995okapi} top-100 BRIGHT candidate pools directly rather than rebuilding them, so that all compared methods rerank the same candidates for a fair comparison (Appendix~\ref{sec:appendix-results-retrieval}). No teacher context is supplied at evaluation.

\vspace{-0.75em}
\paragraph{Table decoding.} Greedy decoding (temperature 0, max 8{,}192 output tokens). The output parser extracts a JSON ranking; unparseable outputs receive nDCG@10 of zero. The parser and decoding budget are identical across all four conditions, so the reported nDCG is an end-to-end measure combining output validity and ranking quality. Parsing failure rates are reported separately in Section~\ref{sec:results-table}. Section~\ref{sec:results-table} also reports an ablation that doubles this budget to 16{,}384 tokens for \baselinename{} only.

\vspace{-0.75em}
\paragraph{Controlled comparison.} The \methodname{} and \baselinename{} checkpoints compared for each model come from the same optimization configuration and the same training step. The methods differ in their serialized context layout and supervised token set (Section~\ref{sec:selective-supervision}). Decoding is greedy (temperature 0) and deterministic. All training FLOPs are $6ND$ estimates, which do not account for LoRA updating only a subset of parameters and so overestimate the compute LoRA training actually uses.

\vspace{-0.75em}
\section{Passage Reranking}
\label{sec:results-passage}
\vspace{-0.75em}

Table~\ref{tab:bright-scale-main} reports BRIGHT nDCG@10 for four models spanning two architecture families.

\begin{table}[t]
\centering
\small
\caption{\textbf{BRIGHT average nDCG@10 ($\times 100$), across model families and scales.} \methodname{} improves over \baselinename{} at every scale and across both families.}
\label{tab:bright-scale-main}
\begin{tabular}{lrrrr}
\toprule
\textbf{Student} & \textbf{\baselinename{}} & \textbf{\methodname{}} & \textbf{Gain} & \textbf{Wins} \\
\midrule
Qwen3-4B  & 29.3 & \textbf{31.5} & +2.2 & 9/12 \\
Qwen3-8B  & 31.5 & \textbf{34.2} & +2.7 & 11/12 \\
GLM-Z1-9B & 30.1 & \textbf{33.2} & +3.1 & 11/12 \\
Qwen3-14B & 32.5 & \textbf{34.7} & +2.2 & 10/12 \\
\bottomrule
\end{tabular}
\vspace{-0.75em}
\end{table}

\begin{figure}[t]
\centering
\includegraphics[width=0.82\textwidth]{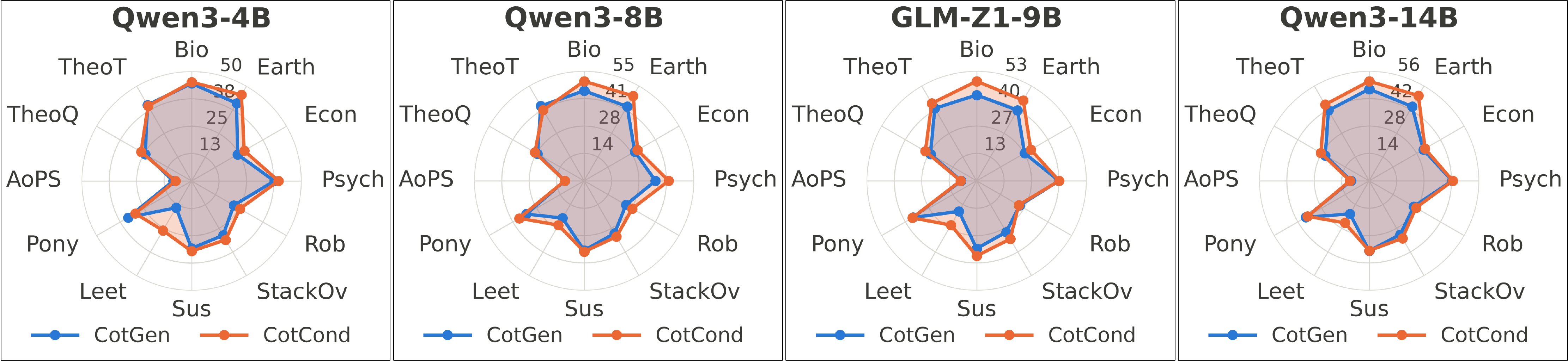}
\caption{\textbf{BRIGHT per-domain nDCG@10, all four model tiers.} \methodname{} (orange) vs.\ \baselinename{} (blue). The losing domains shift from tier to tier, indicating no systematic weakness.}
\label{fig:radar-bright-all}
\vspace{-0.75em}
\end{figure}

\Methodname{} leads at every scale, with absolute gains of 2.2--3.1 points, and holds across the two architecture families we test. Across all four checkpoints and twelve BRIGHT domains, \methodname{} improves in 41 of 48 comparisons. Figure~\ref{fig:radar-bright-all} shows the per-domain profile at each tier: the \methodname{} polygon sits at or outside the \baselinename{} polygon on nearly every axis. Each tier has one to three losing domains, and the sets only partially overlap across tiers.
\paragraph{Per-domain breakdown.} Table~\ref{tab:bright-domains} gives the full twelve-domain breakdown for Qwen3-8B, including an ablation, \maskonlyname{}, that isolates the prompt-side CoT hint: it applies the same masked-loss training as \methodname{} but without ever placing the hint in the prompt, so the full teacher trace sits in the masked assistant-side context instead. \Maskonlyname{} trails \baselinename{} on average (30.33 vs.\ 31.51) and \methodname{} on eleven of twelve domains, so masking the reasoning loss alone does not reproduce \methodname{}'s gains; the hint is doing the work. \Methodname{} improves on eleven of twelve domains, with its largest gains on domains that require reasoning across diverse evidence types (Earth Science, Psychology) or algorithmic structure (LeetCode, Pony) rather than lexical overlap; the single regression, TheoremQA-Theorems ($-2.4$), is a mathematical-proof domain. Regressions are few and shift across model tiers rather than concentrating on a fixed weakness (Appendix~\ref{sec:appendix-perdomain}).


\begin{table}[t]
\centering
\small
\caption{\textbf{BRIGHT, all twelve domains, Qwen3-8B} ($100\times\mathrm{nDCG@10}$). See Appendix~\ref{sec:appendix-table-notes} for the \maskonlyname{} ablation.}
\label{tab:bright-domains}
\begin{tabular}{lcccc}
\toprule
\textbf{Domain} & \textbf{\baselinename{}} & \textbf{\maskonlyname{}} & \textbf{\methodname{}} & \textbf{$\Delta$} \\
\midrule
Biology              & 45.25 & 44.95 & \textbf{50.00} & \dpos{$+4.75$} \\
Earth Science        & 43.13 & 41.46 & \textbf{49.25} & \dpos{$+6.12$} \\
Economics            & 29.27 & 26.63    & \textbf{30.88} & \dpos{$+1.61$} \\
Psychology           & 35.77 & 35.06    & \textbf{42.42} & \dpos{$+6.65$} \\
Robotics             & 24.18 & 25.26    & \textbf{27.87} & \dpos{$+3.69$} \\
StackOverflow        & 30.53 & 29.15    & \textbf{32.41} & \dpos{$+1.88$} \\
Sustainable Living   & 34.92 & 31.92    & \textbf{35.59} & \dpos{$+0.67$} \\
Pony                 & 33.45 & 34.83    & \textbf{37.60} & \dpos{$+4.15$} \\
LeetCode             & 21.59 & 19.92    & \textbf{25.70} & \dpos{$+4.11$} \\
AoPS                 &  9.66 & 8.28    & \textbf{9.70}  & \dpos{$+0.04$} \\
TheoremQA-Theorems   & \textbf{43.37} & 41.44 & 40.95 & \dneg{$-2.42$} \\
TheoremQA-Questions  & 26.96 & 25.02    & \textbf{28.46} & \dpos{$+1.50$} \\
\midrule
\textbf{Average}     & 31.51 & 30.33    & \textbf{34.24} & \dpos{$+2.73$} \\
\bottomrule
\end{tabular}
\vspace{-1.0em}
\end{table}

\vspace{-1.0em}
\section{Table Reranking}
\label{sec:results-table}
\vspace{-1.0em}

Because the motivation for this work is generalization, the Out-of-Domain (OOD) evaluation is the most informative experiment. Table~\ref{tab:table-reranking} evaluates four benchmarks absent from the fine-tuning corpus. All four conditions share the Qwen3-8B backbone and evaluation protocol, differing only in training.

\begin{table}[t]
\centering
\small
\setlength{\tabcolsep}{4pt}
\caption{\textbf{Table reranking results} (nDCG@10, scaled to 0--100) on in-domain and out-of-domain (OOD) benchmarks. See Appendix~\ref{sec:appendix-table-notes} for column conventions and Appendix~\ref{sec:appendix-datasets} for dataset details.}
\label{tab:table-reranking}
\begin{tabular}{llrrrrrrr}
\toprule
& \textbf{Benchmark} & \textbf{Base} & \textbf{\naivename{}} & \textbf{\baselinename{}} & \textbf{\methodname{}} & \textbf{$\Delta$} & \textbf{Unp.\textsubscript{\baselinename{}}} & \textbf{Unp.\textsubscript{\methodname{}}} \\
\midrule
\multirow{4}{*}{\rotatebox[origin=c]{90}{\textit{In-domain}}}
& SQA            & 69.51 & 71.41             & \underline{72.05} & \textbf{73.81}    & \dpos{+1.76} & 0.0  & 0.0  \\
& TAT-QA         & 50.51 & 49.59             & \underline{52.03} & \textbf{54.59}    & \dpos{+2.56} & 6.4  & 0.0  \\
& TabFact        & 60.03 & 62.20             & \underline{66.61} & \textbf{68.90}    & \dpos{+2.29} & 3.9  & 0.3  \\
& NQ-Tables      & 72.10 & 68.88             & \textbf{74.73}    & \underline{72.20} & \dneg{-2.53} & 2.8  & 2.3  \\
\midrule
\multirow{4}{*}{\rotatebox[origin=c]{90}{\textit{OOD}}}
& AITQA          & 77.04 & 68.23             & \textbf{88.25}    & \underline{85.99} & \dneg{-2.26} & 2.3  & 2.5  \\
& FeTaQA         & 89.63 & \underline{91.42} & 88.06             & \textbf{91.95}    & \dpos{+3.89} & 8.3  & 0.0  \\
& OTT-QA (small) & 81.27 & \underline{88.25} & 83.20             & \textbf{90.27}    & \dpos{+7.07} & 14.5 & 0.1  \\
& OpenWikiTables & 88.68 & \underline{91.98} & 90.29             & \textbf{93.78}    & \dpos{+3.49} & 8.6  & 0.1  \\
\bottomrule
\end{tabular}
\vspace{-0.75em}
\end{table}

The four OOD benchmarks each shift away from training in a different way: OpenWikiTable keeps single-table questions like NQ-Tables but requires aggregation and comparison across rows rather than single-cell lookup; FeTaQA keeps single tables but requires a free-form generated answer rather than an extracted one; AITQA keeps a similar QA format but moves to airline-industry tables with a complex, hierarchical structure. OTT-QA is built on top of the training split of HybridQA, and we evaluate on its own test split; the reasoning-operation/answer-format-shift story below is more safely established from FeTaQA and OpenWikiTables alone, with OTT-QA as corroborating evidence. \Methodname{} leads \baselinename{} on the three benchmarks that change the reasoning operation or answer format (FeTaQA, OTT-QA, OpenWikiTables) and trails only on AITQA, where the shift is in table structure rather than task type.

This pattern extends beyond \methodname{} itself: \baselinename{} generalizes poorly enough that \naivename{}, trained without any teacher reasoning, beats it on the same three benchmarks, and on FeTaQA \baselinename{} even falls below the untrained base model. Imitating the teacher's trace, then, does not just fail to help once the reasoning operation or answer format changes, it can actively hurt. \Methodname{} still beats \naivename{} on every OOD benchmark, so dropping the trace is not the fix either; retaining teacher reasoning as context without training the student to reproduce it is what generalizes best here, consistent with the broader pattern reported for CoT fine-tuning \cite{lobo-etal-2025-impact}.

\vspace{-0.75em}
\paragraph{Output validity.} Under identical decoding budgets, \baselinename{} fails to produce a parseable ranking on 0--14.5\% of queries, while \methodname{} fails on 0--2.5\% (Table~\ref{tab:table-reranking}). The gap is widest on OTT-QA (14.5\% vs.\ 0.1\%) and FeTaQA (8.3\% vs.\ 0.0\%). Because unparseable outputs receive a score of zero, the end-to-end nDCG gains in Table~\ref{tab:table-reranking} reflect both ranking quality and output reliability. We attribute this to the supervision signal: under \baselinename{}, the student spends training capacity reproducing the teacher's verbose reasoning patterns, which can interfere with structured output formatting on unseen benchmarks. Under \methodname{}, the supervised tokens are exclusively the well-formatted answer, so the model's capacity concentrates on producing valid output.


\vspace{-0.75em}
\paragraph{Generalization gap.} \Methodname{} wins three of four in-domain benchmarks and three of four OOD benchmarks, but the effect sizes differ sharply: the mean gain over \baselinename{} is $\bar\Delta_{\mathrm{OOD}} = 3.05$ out of domain versus $\bar\Delta_{\mathrm{ID}} = 1.02$ in domain, about three times larger, which would be surprising if the advantage came only from better optimization or a more favorable loss landscape. Part of the gap is mechanical: \baselinename{}'s unparseable-output rate is higher OOD than in domain (Table~\ref{tab:table-reranking}) and every such failure scores zero, so some of the larger OOD margin reflects output-format reliability rather than ranking quality alone. We hypothesize trace imitation also binds the student to training-distribution surface patterns that transfer poorly, though we cannot isolate this from the reliability effect here. The single in-domain regression (NQ-Tables, $-2.53$) does not break the overall pattern.

\begin{table}[t]
\centering
\small
\setlength{\tabcolsep}{4pt}
\caption{\textbf{Truncation-budget ablation} (in-domain, nDCG@10, scaled to 0--100). See Appendix~\ref{sec:appendix-table-notes} for the ablation setup and column conventions.}
\label{tab:compute-matched}
\begin{tabular}{lrrrr}
\toprule
\textbf{Condition} & \textbf{SQA} & \textbf{TAT-QA} & \textbf{TabFact} & \textbf{NQ-Tables} \\
\midrule
\baselinename{} (8{,}192 tok.)           & 72.05          & 52.03          & 66.61          & 74.73 \\
\baselinename{} (extended, 16{,}384 tok.) & \textbf{75.00} & 53.50          & 65.90          & \textbf{77.30} \\
\methodname{}                             & 73.80          & \textbf{54.60} & \textbf{68.90} & 72.20 \\
\midrule
$\Delta$ (\methodname{} $-$ extended)     & \dneg{-1.20}   & \dpos{+1.10}   & \dpos{+3.00}   & \dneg{-5.10} \\
\bottomrule
\end{tabular}
\vspace{-1.0em}
\end{table}

\vspace{-0.75em}
\paragraph{Is \methodname{}'s in-domain shortfall a truncation artifact?} A natural alternative explanation for \methodname{}'s two in-domain losses is decoding budget: if \baselinename{} frequently exhausts its 8{,}192-token budget mid-trace, its answer would be based on an incomplete rationale, and \methodname{} could look better for that reason alone. We test this by re-decoding \baselinename{} with the output budget doubled to 16{,}384 tokens, changing nothing else about the checkpoint or decoding procedure (Table~\ref{tab:compute-matched}). This ablation covers the four in-domain benchmarks only, and extends the budget for \baselinename{} alone.

Doubling the budget helps \baselinename{} on SQA and NQ-Tables, where it now leads \methodname{} by 1.2 and 5.1 points, and leaves \methodname{} ahead only on TAT-QA (+1.1) and TabFact (+3.0). Averaged over the four benchmarks, extended-budget \baselinename{} beats \methodname{} by 0.55 points. So at least part of \methodname{}'s in-domain shortfall on SQA and NQ-Tables is consistent with \baselinename{} being cut off before it finishes reasoning: given enough room to think, \baselinename{} matches or exceeds \methodname{} in domain.

\vspace{-1em}
\section{Efficiency}
\label{sec:efficiency}
\vspace{-1em}

Table~\ref{tab:efficiency} reports training FLOPs (via the standard $6ND$ estimator) and inference wall-clock time for every passage-reranking model pair.

\begin{table}[t]
\centering
\small
\setlength{\tabcolsep}{4pt}
\caption{\textbf{Training FLOPs and inference time, passage reranking.} Training values are $6ND$ estimates in $10^{18}$ FLOPs. Inference is wall-clock minutes on a single A100 with vLLM, greedy decoding, batch size 1. \methodname{} uses less compute and is faster at every scale --- an accuracy--efficiency Pareto improvement.}
\label{tab:efficiency}
\begin{tabular}{lrrrrrr}
\toprule
& \multicolumn{3}{c}{\textbf{Training ($\times 10^{18}$ FLOPs)}} & \multicolumn{3}{c}{\textbf{Inference (min)}} \\
\cmidrule(lr){2-4}\cmidrule(lr){5-7}
\textbf{Student} & \textbf{\baselinename{}} & \textbf{\methodname{}} & \textbf{Red.} & \textbf{\baselinename{}} & \textbf{\methodname{}} & \textbf{Red.} \\
\midrule
Qwen3-4B  & 10.93 & 7.55 & 30.9\% & 118.5 & 79.8 & 32.7\% \\
Qwen3-8B  & 21.50 & 14.70 & 31.6\% & 132.0 & 95.5 & 27.7\% \\
GLM-Z1-9B & 23.70 & 16.61 & 29.9\% & 104.4 & 88.3 & 15.4\% \\
Qwen3-14B & 35.29 & 26.89 & 23.8\% & 176.1 & 139.0 & 21.1\% \\
\bottomrule
\end{tabular}
\vspace{-1.0em}
\end{table}

\vspace{-0.75em}
\paragraph{Training compute.} \Methodname{} reduces estimated training FLOPs by 24--32\% on passage reranking. The reduction comes from shorter realized training sequences: the lexical filter (Section~\ref{sec:reasoning-transformation}) removes verdict-bearing sentences from the reasoning hint before it enters the prompt, so \methodname{}'s total sequence, prompt plus response, contains fewer tokens than \baselinename{}'s unfiltered trace of the same passage pair. This is a genuine reduction in processed tokens, not an artifact of where the sequence cutoff falls, since masking labels alone does not eliminate transformer computation. Step counts are identical across methods.

\vspace{-0.75em}
\paragraph{Inference speed.} Inference is 15--33\% faster at every scale, measured on a single A100 GPU with vLLM \cite{kwon2023efficient}. The speedup is consistent with shorter generated reasoning: on 899 matched Qwen3-8B passage pairs, \methodname{} generates a mean of 198 words per trace versus 312 for \baselinename{}, a $1.58\times$ reduction holding across all twelve BRIGHT domains (Appendix~\ref{sec:appendix-generated_output} gives the per-domain range and the analogous $1.5$--$1.8\times$ reduction on table reranking).
The length reduction follows from the changed supervision. Under \baselinename{}, the student is trained to reproduce the teacher's often-verbose reasoning, and this verbosity persists at inference even when the teacher context is absent. Under \methodname{}, the student is never rewarded for generating long reasoning traces; it learns to reason only as much as needed to produce a correct answer. Combined with the accuracy gains, these efficiency improvements yield an accuracy--efficiency Pareto improvement at every tested scale.

\vspace{-0.5em}
\section{What Changes in the Student's Reasoning}
\label{sec:analysis}
\vspace{-0.5em}

We now examine what removing trace supervision changes about the student's generated reasoning. If \methodname{} degraded reasoning quality, we would expect its traces to be less structured or less connected to its outputs. We test this by analyzing Qwen3-8B traces on matched inputs across training conditions, using the same lexical extractors for both methods.

\vspace{-0.5em}
\paragraph{Verdict Timing and Stability. }\label{sec:analysis-commitment} A lexical detector identifies relevance verdicts (``relevant,'' ``irrelevant,'' \texttt{true}/\texttt{false}) in 899 matched passage-reranking pairs. Table~\ref{tab:commitment} reports when the first verdict appears, how many times it is restated (avg.\ mentions), and how often the model reverses its stated judgment ($\geq$1 reversal).

\begin{table}[t]
\centering
\small
\setlength{\tabcolsep}{4pt}
\caption{\textbf{Verdict timing and reversals} for Qwen3-8B on 899 matched passage-reranking pairs. Position is the fraction of the reasoning trace elapsed at the first detected relevance verdict (0\,=\,start, 1\,=\,end).}
\label{tab:commitment}
\begin{tabular}{lrrrrrr}
\toprule
& \textbf{Detection} & \textbf{Mean first} & \textbf{Median first} & \textbf{Q1} & \textbf{$\geq$1 reversal} & \textbf{Avg.} \\
\textbf{Method} & \textbf{(\% traces)} & \textbf{position} & \textbf{position} & \textbf{position} & \textbf{(\% traces)} & \textbf{mentions} \\
\midrule
\baselinename{} & 98.6 & 0.44 & 0.37 & 0.05 & 16.0 & 2.3 \\
\methodname{} & 89.4 & 0.76 & 0.85 & 0.66 & 5.2 & 1.4 \\
\bottomrule
\end{tabular}
\vspace{-1.0em}
\end{table}

\Methodname{} students state their first verbalized verdict later (median position 0.85 vs.\ 0.37, Mann--Whitney $p=1.5\times10^{-69}$, holding across all twelve BRIGHT domains), reverse it less often (5.2\% vs.\ 16.0\%), and restate it less (1.4 mentions per trace vs.\ 2.3). The gap widens at the 25th percentile (Q1\,=\,0.05 for \baselinename{} vs.\ 0.66 for \methodname{}), and \baselinename{} detects a verdict in more traces overall (98.6\% vs.\ 89.4\%), suggesting some \methodname{} traces express judgment only implicitly through the final answer token.

This is consistent with \baselinename{} imitating the teacher's early-preliminary-judgment-then-elaboration pattern, sometimes contradicting itself when later evidence conflicts with that early commitment, whereas \methodname{} has no such pattern to copy and instead gathers evidence before stating a judgment.

\vspace{-1em}
\paragraph{Trace-Answer Association. }\label{sec:analysis-association} For table reranking, we extract two features from each reasoning trace, \emph{mentions} of ``Table~$k$'' and the \emph{recency} of its last mention. We combine them into a joint feature $F$ and measure normalized mutual information with the model's top-ranked choice $\hat Y$ and the gold label $Y$ over 132{,}000 candidate-level observations.

\begin{table}[t]
\centering
\small
\setlength{\tabcolsep}{4pt}
\caption{\textbf{Trace--answer association} for Qwen3-8B table reranking over 132k candidate-level observations. Each cell is $I(\cdot\,;\,\text{target})/H(\text{target})$; higher values indicate stronger predictive association.}
\label{tab:trace-association}
\begin{tabular}{lrrr|r}
\toprule
& \multicolumn{3}{c|}{\textbf{Trace $\to$ model's choice} $\hat Y$} & \textbf{Trace $\to$ gold} $Y$ \\
\cmidrule(lr){2-4} \cmidrule(lr){5-5}
\textbf{Condition} & \textbf{Mentions} & \textbf{Recency} & \textbf{Both ($F$)} & \textbf{Both ($F$)} \\
\midrule
Base & 0.313 & 0.311 & 0.373 & 0.140 \\
\naivename{} & 0.278 & 0.163 & 0.317 & 0.097 \\
\methodname{} & 0.214 & 0.167 & 0.264 & 0.100 \\
\baselinename{} & 0.111 & 0.101 & 0.144 & 0.049 \\
\bottomrule
\end{tabular}
\vspace{-1.0em}
\end{table}

\Methodname{} shows substantially stronger trace--answer association than \baselinename{} for both its own predictions ($I(F;\hat Y)/H=0.264$ vs.\ $0.144$) and the gold label ($0.100$ vs.\ $0.049$); \baselinename{} in fact exhibits the weakest association among all four conditions despite directly supervising the teacher's reasoning trace. Removing direct trace supervision therefore does not reduce the measured connection between generated reasoning and final predictions, though these associations remain observational and do not establish causal faithfulness. Further analysis of surface style and voice is provided in Appendix~\ref{sec:analysis-style}.
 
\vspace{-1.0em}
\section{Conclusion}
\label{sec:conclusion}
\vspace{-1.0em}

\Methodname{} studies whether teacher reasoning can serve as training-time context rather than a direct imitation target for reranking. Retaining teacher reasoning while supervising only the final answer improves performance in most evaluated settings, with particularly strong gains on held-out benchmarks (exceptions: AITQA, NQ-Tables, and the Table~\ref{tab:compute-matched} ablation, where a larger decoding budget lets \baselinename{} match or exceed \methodname{} in domain), while also producing fewer malformed outputs and shorter generations at no cost of student rollouts or iterative teacher feedback. Trace-level analyses further show \methodname{} students state verdicts later, reverse them less often, and generate reasoning more predictive of their final decisions than \baselinename{}'s, though still less predictive than the base model's. These gains are task-dependent: exploratory mathematical-reasoning results show the opposite pattern (Appendix~\ref{sec:appendix-generative}), and broader validation across tasks and training formulations remains an important direction.
\vspace{-1.0em}
\section{Limitations}
\label{sec:limitations}
\vspace{-1.0em}

As with any empirical study, our conclusions are drawn from a finite set of models, datasets, and hyperparameter choices, and may not hold under every configuration outside the ones we test. Reported metrics and efficiency estimates rely on standard approximations and measurement conventions common in the literature, which can differ from other studies' methodology. Broader replication across additional tasks, model families, and training regimes would further establish the generality of the trends we observe. Observed associations between measured quantities do not by themselves establish the underlying causal mechanism, and results at the scales and compute budgets studied here may not extrapolate uniformly to substantially larger or smaller settings. Some implementation and design choices were fixed rather than exhaustively tuned, and a different choice could shift absolute numbers without necessarily changing the overall trend.
\subsection*{AI use statement}

We used generative AI tools for interpreting experimental results and for generating synthetic data: the teacher reasoning traces underlying our fine-tuning corpora were produced by a large language model (DeepSeek-R1 for table reranking; the public Rank1 training corpus for passage reranking), and we used a general-purpose AI assistant to help identify and characterize patterns in our experimental results during analysis. We have not used generative AI tools to design the \methodname{} methodology, our experiments, or the training/evaluation code; the training and evaluation pipelines were implemented by the authors without AI assistance. We additionally used generative AI tools for drafting and editing the text and github repository for this paper. All AI-assisted analysis and writing was reviewed by the authors, who verified reported numbers against the underlying experimental logs and take full responsibility for the final content of this work, including all text, claims, and reported results.

\subsection*{Ethics statement}

This work uses only publicly released models (Qwen3, GLM-Z1, DeepSeek-R1) and publicly available academic benchmarks (BRIGHT, SQA, TAT-QA, TabFact, NQ-Tables, HybridQA/MultiTableQA, AITQA, FeTaQA, OTT-QA, OpenWikiTables), each used under the terms of its original release and cited accordingly (Appendix~\ref{sec:appendix-datasets}). No new human-subject data was collected, and none of the datasets involve personally identifiable information. We do not anticipate specific harms beyond the general dual-use considerations inherent to information retrieval and reranking research.

\subsection*{Reproducibility statement}

Our full training and evaluation configuration is in Section~\ref{sec:experimental-setup} and Appendix~\ref{sec:appendix-method-details}. For reproducibility, we release our code, results, and models. 

\bibliography{EKLAV}
\bibliographystyle{iclr2027_conference}

\appendix
\section{Method Details}
\label{sec:appendix-method-details}

This section records the preprocessing, masking, and deployment details of the training recipe.

\subsection{Hint construction}
\label{sec:appendix-hint-details}

\paragraph{Sentence segmentation and split.} The pointwise trace is split into sentences using sentence-boundary detection. If the trace contains $m$ sentences, the split is at $k=\lfloor m/2\rfloor$. The first part is filtered and placed in the prompt, and the remaining teacher-written sentences are retained as masked assistant-side context. A midpoint by sentence count need not be a midpoint by token count. No split-ratio ablation is reported.

\paragraph{Lexical judgment filter.} The recorded filter removes verdict-bearing sentences from the prompt prefix, keeping the prompt-side hint free of the teacher's verdict. Its patterns include relevance vocabulary and direct assertions such as ``is true,'' ``is false,'' and ``the answer is.'' The description also records a heuristic based on following prepositions to distinguish task framing from a judgment. The exact executable patterns are needed to reproduce the transformation.

If filtering removes every sentence from the prefix, the hint is left empty rather than falling back to an unfiltered sentence, so no verdict-bearing text reaches the prompt through this path. The remaining teacher continuation is also available during training, and the full listwise trace is placed in the prompt. Consequently, the method should not be described as eliminating answer information from training context. No teacher-derived hint or continuation is provided at evaluation.

For the exploratory mathematical setting, the recorded filter additionally matches boxed answers and phrases such as ``the solution is.'' The provided summary does not include a separate audit of its accuracy.

\subsection{Sample prompts}
\label{sec:appendix-sample-prompts}

We show the base prompt structure used for each task, as issued at evaluation (no teacher-derived hint or trace is present at evaluation for either task; Section~\ref{sec:masking-mechanics}). At training time, \methodname{} inserts the filtered reasoning hint $h_\tau(r)$ into the pointwise prompt (Section~\ref{sec:reasoning-transformation}) or the complete teacher trace $r$ into the listwise prompt (Section~\ref{sec:reasoning-transformation}); we omit example hint/trace text here since it is produced by an external data-generation pipeline not included with this submission.

\paragraph{Pointwise passage reranking.}
\begin{promptbox}
\textbf{System:} You are a careful retrieval assistant that judges whether a passage is relevant to a user's query.\\[4pt]
\textbf{User:} Determine if the following passage is relevant to the query. Answer only with `true' or `false'.\\[2pt]
Query: \{query\}\\
Passage: \{passage\}
\end{promptbox}
The student's response begins with a \texttt{<think>...</think>} reasoning span followed by \texttt{true} or \texttt{false}; only the final verdict token is scored, $s(q,d)=p_\theta(\texttt{true}\mid q,d,\hat r)$ (Section~\ref{sec:experimental-setup}).

\paragraph{Listwise table reranking.}
\begin{promptbox}
\textbf{System:} You are a careful retrieval assistant that ranks candidate tables by their relevance to a user's question.\\[4pt]
\textbf{User:} Question: \{question\}\\[2pt]
Candidate tables to rank:\\[2pt]
\#\#\# Table 1\\
\{table\_1\}\\[2pt]
\#\#\# Table 2\\
\{table\_2\}\\
...\\[4pt]
Reason through each table's relevance step-by-step.\\
Then output exactly: \{"ranked\_tables": [1, 2, 3, ...]\}
\end{promptbox}
Candidate tables are windowed (at most 20 tables or 20{,}000 prompt tokens per window, stride 10) and the model's per-window ranking is merged into a single final ranking over the full candidate pool (Appendix~\ref{sec:appendix-datasets}).

\subsection{Masking auto-inserted reasoning blocks}
\label{sec:appendix-masking}

Some chat templates insert an empty reasoning block into a response that lacks one. Under an answer-only objective, the inserted block must also be masked; otherwise those tokens receive direct next-token supervision. The recorded implementation encountered a boundary-matching issue when tokenizing the closing tag in isolation, because subword tokenization can depend on adjacent characters.

The implementation instead locates the closing boundary in the decoded response and maps the character offset to the smallest token index whose decoded prefix covers the tag. The search is restricted to the assistant response, because the prompt may also mention reasoning delimiters. This behavior still requires verification for each template, including merged boundary tokens and missing or repeated tags. Mask positions must also be aligned with the model's next-token label shift.

\subsection{Gradient flow through teacher-forced context}
\label{sec:appendix-masking-mechanics}

Let $z_t^{(\ell)}$ denote an intermediate representation of a teacher-supplied reasoning token at position $t$. A masked token has no direct next-token loss. Its representation can nevertheless influence later answer positions through subsequent attention layers, so $\partial\mathcal{L}_A/\partial z_t^{(\ell)}$ can be nonzero along those paths. Loss masking therefore does not delete reasoning context or block all gradients through its processing.

This answer-loss pathway exists in both training conditions. \Baselinename{} additionally supervises prediction of the teacher reasoning tokens. All tokens supplied as conditioning context during training are teacher-forced, including \methodname{}'s assistant-side continuation. The student computes their representations; it does not generate those tokens as a training rollout. Gradients update parameters through the processing of fixed context and do not differentiate through sampled discrete reasoning. This observation alone establishes neither causal dependence on generated reasoning nor faithfulness at inference.

\subsection{Training and deployment context}
\label{sec:appendix-deployment}

Before sequence truncation, the teacher-derived context is
\begin{equation}
C_T=\bigl(h_\tau(r),c_\tau(r)\bigr).
\end{equation}
For pointwise training, this is $(F(r_{1:k}),r_{k+1:m})$; for listwise training, it is $(r,\varnothing)$. The full teacher trace is required before filtering and serialization. The context retained in the model input is subject to the task's sequence cutoff.

At evaluation, the model receives the task input without teacher reasoning. It generates its own reasoning before answering,
\begin{equation}
\hat R_S\sim p_\theta(\cdot\mid x),\qquad C_T\longrightarrow\hat R_S.
\label{eq:context-shift}
\end{equation}
This is the reasoning context shift discussed in Section~\ref{sec:method}. Evaluation also removes the prompt-side framing used to supply teacher context. All reported test results are obtained in this setting. The observed performance therefore measures the trained model after privileged context has been withdrawn, without proving that the teacher's reasoning procedure has been internalized.

\section{Full Results}
\label{sec:appendix}

This appendix collects the reported benchmark tables, cost records, and exploratory summaries, including regressions.

\subsection{Main-text table conventions}
\label{sec:appendix-table-notes}

Table~\ref{tab:bright-domains} (BRIGHT, all twelve domains) reports \maskonlyname{}, an ablation that keeps \methodname{}'s masked-loss training but removes the prompt-side CoT hint, to test whether the hint itself, not just the masking, drives \methodname{}'s gains (Section~\ref{sec:results-passage}).

Table~\ref{tab:table-reranking} (table reranking results) bolds the highest score among the four conditions per row and underlines the second highest; $\Delta$ is the absolute gain of \methodname{} over \baselinename{}, and the \textit{Unp.} columns report the unparseable-output rate (\%) for \baselinename{} and \methodname{} under the same parser and decoding budget, with such failures scoring zero in the nDCG columns. See Appendix~\ref{sec:appendix-datasets} for the in-domain and OOD datasets.

Table~\ref{tab:compute-matched} (truncation-budget ablation) doubles \baselinename{}'s normal 8{,}192-token output budget to 16{,}384 tokens, the only change, to test whether \methodname{}'s gains simply come from cases where \baselinename{} was cut off mid-trace; bold marks the highest score per benchmark and $\Delta$ is \methodname{} minus \baselinename{} (extended).

\subsection{Table reranking datasets}
\label{sec:appendix-datasets}

This paper uses eight datasets for table reranking: four in-domain benchmarks drawn from two prior benchmark papers, NQ-Tables \cite{nq_tables} and MultiTableQA \cite{multiTableQA_dataset}, and four further datasets used exclusively for out-of-domain evaluation.

\paragraph{Training (in-domain).} We use the NQ-Tables \cite{nq_tables} training set for single-table, correct-type questions. For multi-table QA, we use only the HybridQA \cite{chen-etal-2020-hybridqa} portion of the MultiTableQA benchmark \cite{multiTableQA_dataset}, and only its training split. We keep the remaining datasets from the MultiTableQA benchmark for evaluation and still treat them as in-domain.

\paragraph{Testing (out-of-domain).} For out-of-domain (OOD) evaluation, we draw on the IBM Research Table-Text IR Evaluation collection \cite{doshi2026tableir} and report results on AIT-QA \cite{katsis2021aitqa}, FeTaQA \cite{nan2022fetaqa}, OTT-QA \cite{chen2021ottqa}, and Open-WikiTable \cite{kweon-etal-2023-open} (Table~\ref{tab:table-reranking}); none of these datasets' examples appear in the fine-tuning corpus. OTT-QA is built on top of HybridQA's training split, and we evaluate on OTT-QA's own test split (Section~\ref{sec:results-table}).

\paragraph{Candidate pool construction.} Both training and evaluation candidate pools are built with the same retrieval pipeline: BM25 \cite{robertson1995okapi}, SPLADE-v3, and all-mpnet-base-v2 each retrieve a top-100 pool per query, and the three rankings are merged by reciprocal rank fusion (RRF, $k=60$) into a single top-100 pool, with any missing gold table injected. Combining a lexical retriever (BM25) with two dense retrievers (SPLADE-v3, all-mpnet-base-v2) ensures candidates are not selected by semantic similarity alone, following the retrieval setup used by Rank1 \cite{wellerRank1TesttimeCompute2025} for passage reranking.

\subsection{Passage reranking: per-domain breakdown}
\label{sec:appendix-perdomain}

The Qwen3-8B per-domain breakdown is given in the main text (Table~\ref{tab:bright-domains}), where the largest gains are on Earth Science (+6.1) and Psychology (+6.7), domains that require reasoning across diverse evidence types rather than lexical overlap, and substantial gains also appear on the coding domains LeetCode (+4.1) and Pony (+4.2). The single regression, TheoremQA-Theorems ($-2.4$), is a mathematical-proof domain where the teacher trace may carry procedural knowledge (proof structure, theorem application) that benefits from direct imitation.

The remaining three model tiers show similar profiles: \methodname{} leads on the average and on the majority of domains. Regression domains only partially overlap across tiers: at 4B, the regressions are on Pony, AoPS, and TheoremQA-Theorems; at 9B, only Robotics regresses; at 14B, Pony and Sustainable Living regress. Pony regresses at both 4B and 14B but not at 8B or 9B. Regressions are few (one to three domains per tier) and no single domain regresses at every tier.

\subsection{Passage retrieval: full breakdown with baselines}
\label{sec:appendix-results-retrieval}

We use the candidate pools released by Rank1 \cite{wellerRank1TesttimeCompute2025}: the BM25 \cite{robertson1995okapi} top-100 passages per BRIGHT query, exactly as published, rather than reconstructing them ourselves. All four training conditions rerank this same fixed pool, so any nDCG@10 difference reflects reranking quality rather than a difference in candidate recall. Our pointwise scoring formulation, generating a reasoning trace before a \texttt{true}/\texttt{false} verdict, is likewise inspired by Rank1's approach to test-time compute for reranking.

Table~\ref{tab:bright-main} gives the complete per-domain BRIGHT breakdown including published baselines. Table~\ref{tab:consolidated-results-full} gives the full consolidated view (training FLOPs and inference speed alongside accuracy).

\begin{table}[p]
\centering
\small
\setlength{\tabcolsep}{3pt}
\caption{\textbf{BRIGHT contextual baselines and paired results, StackExchange domains.} Scores are $100\times\mathrm{nDCG@10}$. Published rows provide context \cite{suBRIGHTRealisticChallenging2025,wellerRank1TesttimeCompute2025}; our paired rows share the recipe in Section~\ref{sec:experimental-setup}. Delta rows retain the supplied relative changes and can differ from ratios of rounded cells.}
\label{tab:bright-main}
\begin{tabularx}{\linewidth}{@{}X*{7}{r}@{}}
\toprule
Model & Bio & Earth & Econ & Psych & Rob & StackOv & Sus \\
\midrule
BM25 & 19.2 & 27.1 & 14.9 & 12.5 & 13.5 & 16.5 & 15.2 \\
BM25 on GPT-4o CoT & 53.6 & 53.6 & 24.3 & 38.6 & 18.8 & 22.7 & 25.9 \\
MonoT5-3B & 16.0 & 24.0 & 17.7 & 19.5 & 8.0 & 10.5 & 19.5 \\
RankLLaMA-7B & 17.5 & 15.5 & 13.1 & 13.6 & 17.9 & 6.9 & 16.9 \\
RankLLaMA-13B & 21.6 & 19.1 & 16.3 & 14.0 & 15.7 & 7.7 & 18.5 \\
Rank1-7B & 48.8 & 36.7 & 20.8 & 35.0 & 22.0 & 18.7 & 36.2 \\
Rank1-14B & 49.3 & 37.7 & 22.6 & 35.2 & 22.5 & 20.8 & 33.6 \\
Rank1-32B & 49.7 & 35.8 & 22.0 & 37.5 & 22.5 & 21.7 & 35.0 \\
\midrule
\rowcolor{cotgenrow} Qwen3-4B (\baselinename{}) & 44.9 & 41.3 & 24.4 & 38.0 & 22.4 & 28.9 & 31.0 \\
\rowcolor{cotcondrow} Qwen3-4B (\methodname{}) & 45.5 & 45.8 & 27.9 & 39.9 & 25.6 & 31.3 & 32.3 \\
$\Delta$ vs.\ \baselinename{} & \dpos{+1.4\%} & \dpos{+10.8\%} & \dpos{+14.5\%} & \dpos{+5.0\%} & \dpos{+14.6\%} & \dpos{+8.1\%} & \dpos{+4.3\%} \\
\midrule
\rowcolor{cotgenrow} Qwen3-8B (\baselinename{}) & 45.2 & 43.1 & 29.3 & 35.8 & 24.2 & 30.5 & 34.9 \\
\rowcolor{cotcondrow} Qwen3-8B (\methodname{}) & 50.0 & 49.2 & 30.9 & 42.4 & 27.9 & 32.4 & 35.6 \\
$\Delta$ vs.\ \baselinename{} & \dpos{+10.5\%} & \dpos{+14.2\%} & \dpos{+5.5\%} & \dpos{+18.6\%} & \dpos{+15.3\%} & \dpos{+6.2\%} & \dpos{+1.9\%} \\
\midrule
\rowcolor{cotgenrow} GLM-Z1-9B (\baselinename{}) & 41.8 & 39.7 & 26.9 & 40.0 & 24.1 & 28.8 & 33.0 \\
\rowcolor{cotcondrow} GLM-Z1-9B (\methodname{}) & 48.5 & 45.3 & 30.4 & 40.1 & 23.6 & 32.7 & 36.6 \\
$\Delta$ vs.\ \baselinename{} & \dpos{+16.1\%} & \dpos{+14.2\%} & \dpos{+13.2\%} & \dpos{+0.2\%} & \dneg{$-$2.1\%} & \dpos{+13.5\%} & \dpos{+10.9\%} \\
\midrule
\rowcolor{cotgenrow} Qwen3-14B (\baselinename{}) & 46.6 & 43.6 & 31.7 & 42.1 & 26.1 & 31.5 & 35.6 \\
\rowcolor{cotcondrow} Qwen3-14B (\methodname{}) & 50.6 & 50.0 & 32.7 & 42.4 & 27.4 & 33.7 & 35.5 \\
$\Delta$ vs.\ \baselinename{} & \dpos{+8.6\%} & \dpos{+14.7\%} & \dpos{+3.2\%} & \dpos{+0.8\%} & \dpos{+5.1\%} & \dpos{+7.2\%} & \dneg{$-$0.3\%} \\
\bottomrule
\end{tabularx}
\end{table}

\begin{table}[p]
\centering
\small
\setlength{\tabcolsep}{3pt}
\caption{\textbf{BRIGHT coding and theorem domains, with overall averages.} This continues Table~\ref{tab:bright-main}. Scores use the same scale; the delta rows are the supplied relative percentage changes. The Qwen3-8B AoPS improvement is visible only at higher precision (Table~\ref{tab:bright-domains}).}
\label{tab:bright-main-rest}
\begin{tabularx}{\linewidth}{@{}X*{6}{r}@{}}
\toprule
Model & Leet & Pony & AoPS & TheoQ & TheoT & Avg \\
\midrule
BM25 & 24.4 & 7.9 & 6.0 & 13.0 & 6.9 & 14.8 \\
BM25 on GPT-4o CoT & 19.3 & 17.7 & 3.9 & 18.9 & 20.2 & 26.5 \\
MonoT5-3B & 17.2 & 29.2 & 7.1 & 20.3 & 12.0 & 16.8 \\
RankLLaMA-7B & 8.4 & 46.8 & 2.2 & 4.5 & 3.5 & 13.9 \\
RankLLaMA-13B & 8.8 & 31.1 & 1.7 & 4.4 & 4.9 & 13.7 \\
Rank1-7B & 12.7 & 31.2 & 6.3 & 23.7 & 37.8 & 27.5 \\
Rank1-14B & 17.7 & 33.2 & 8.4 & 22.5 & 41.4 & 28.7 \\
Rank1-32B & 18.8 & 32.5 & 10.8 & 22.9 & 43.7 & 29.4 \\
\midrule
\rowcolor{cotgenrow} Qwen3-4B (\baselinename{}) & 14.2 & 33.6 & 8.4 & 24.6 & 40.4 & 29.3 \\
\rowcolor{cotcondrow} Qwen3-4B (\methodname{}) & 26.3 & 29.9 & 7.4 & 26.8 & 39.9 & 31.5 \\
$\Delta$ vs.\ \baselinename{} & \dpos{+85.9\%} & \dneg{$-$11.0\%} & \dneg{$-$12.6\%} & \dpos{+8.9\%} & \dneg{$-$1.1\%} & \dpos{+7.6\%} \\
\midrule
\rowcolor{cotgenrow} Qwen3-8B (\baselinename{}) & 21.6 & 33.4 & 9.7 & 27.0 & 43.4 & 31.5 \\
\rowcolor{cotcondrow} Qwen3-8B (\methodname{}) & 25.7 & 37.6 & 9.7 & 28.5 & 41.0 & 34.2 \\
$\Delta$ vs.\ \baselinename{} & \dpos{+19.0\%} & \dpos{+12.4\%} & \dpos{+0.3\%} & \dpos{+5.6\%} & \dneg{$-$5.6\%} & \dpos{+8.7\%} \\
\midrule
\rowcolor{cotgenrow} GLM-Z1-9B (\baselinename{}) & 17.2 & 35.8 & 7.5 & 25.9 & 40.7 & 30.1 \\
\rowcolor{cotcondrow} GLM-Z1-9B (\methodname{}) & 24.9 & 36.1 & 7.8 & 28.8 & 43.6 & 33.2 \\
$\Delta$ vs.\ \baselinename{} & \dpos{+45.0\%} & \dpos{+1.0\%} & \dpos{+4.0\%} & \dpos{+11.2\%} & \dpos{+7.2\%} & \dpos{+10.3\%} \\
\midrule
\rowcolor{cotgenrow} Qwen3-14B (\baselinename{}) & 19.4 & 37.1 & 9.2 & 25.7 & 41.3 & 32.5 \\
\rowcolor{cotcondrow} Qwen3-14B (\methodname{}) & 24.7 & 36.1 & 9.7 & 28.3 & 44.7 & 34.7 \\
$\Delta$ vs.\ \baselinename{} & \dpos{+27.6\%} & \dneg{$-$2.7\%} & \dpos{+5.5\%} & \dpos{+10.1\%} & \dpos{+8.2\%} & \dpos{+6.7\%} \\
\bottomrule
\end{tabularx}
\end{table}

\begin{table}[H]
\centering
\caption{\textbf{Consolidated results, full version.} Estimated training FLOPs and inference elapsed time.}
\label{tab:consolidated-results-full}
\resizebox{\textwidth}{!}{%
\begin{tabular}{llccc}
\toprule
Model & Method & BRIGHT avg. & Estimated FLOPs & Inference time \\
\midrule
\multirow{2}{*}{Qwen3-4B} & \baselinename{} & 29.3 & $1.093\times10^{19}$ & 118.5 min (BRIGHT) \\
 & \methodname{} & 31.5 & $7.553\times10^{18}$ & 79.8 min (BRIGHT) \\
\midrule
\multirow{2}{*}{Qwen3-8B} & \baselinename{} & 31.5 & $2.150\times10^{19}$ & 132.0 min (BRIGHT) \\
 & \methodname{} & 34.2 & $1.470\times10^{19}$ & 95.5 min (BRIGHT) \\
\midrule
\multirow{2}{*}{GLM-Z1-9B} & \baselinename{} & 30.1 & $2.370\times10^{19}$ & 104.4 min (BRIGHT) \\
 & \methodname{} & 33.2 & $1.661\times10^{19}$ & 88.3 min (BRIGHT) \\
\midrule
\multirow{2}{*}{Qwen3-14B} & \baselinename{} & 32.5 & $3.529\times10^{19}$ & 176.1 min (BRIGHT) \\
 & \methodname{} & 34.7 & $2.689\times10^{19}$ & 139.0 min (BRIGHT) \\
\bottomrule
\end{tabular}%
}
\end{table}

\begin{figure}[ht]
\centering
\includegraphics[width=\textwidth]{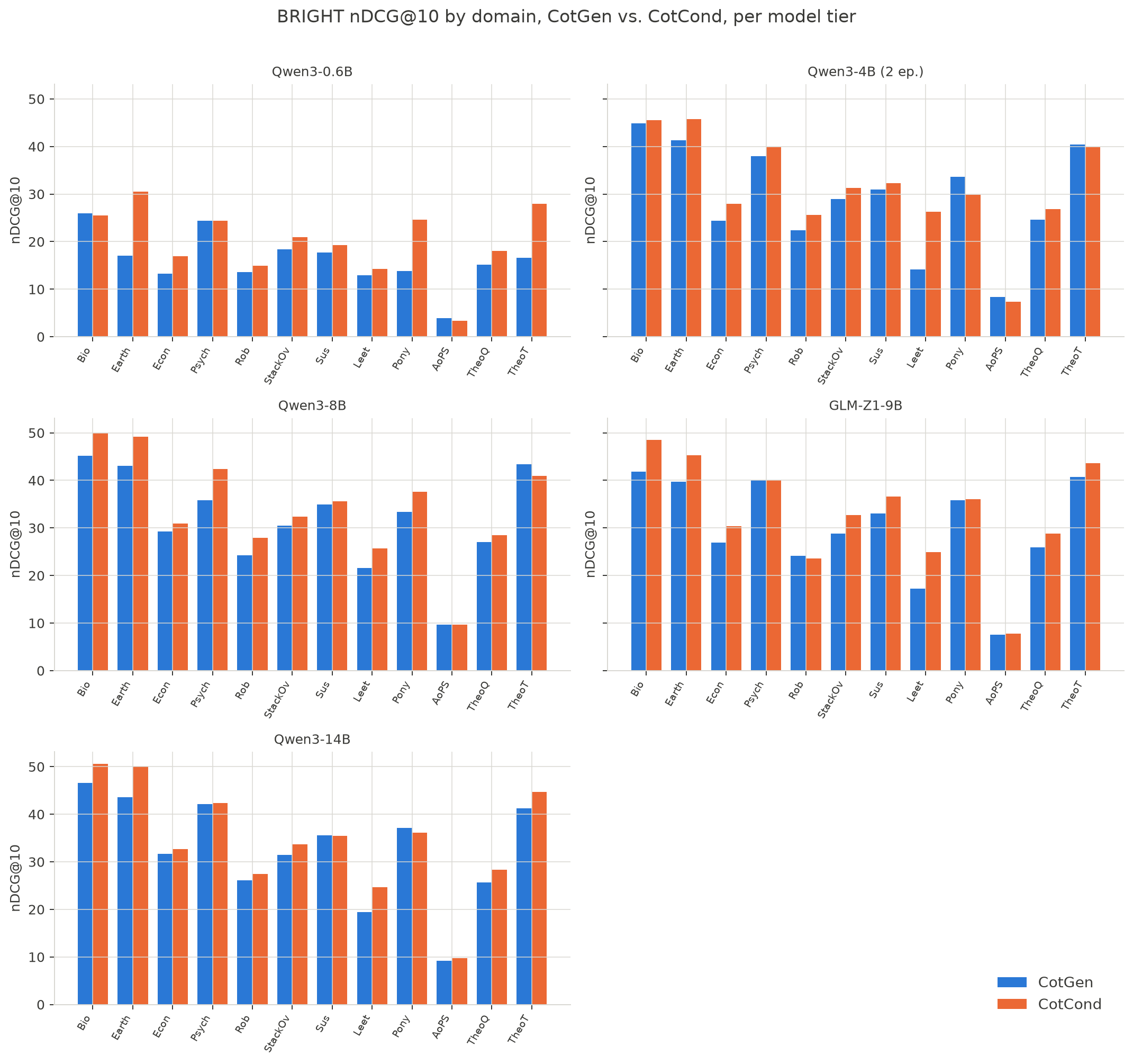}
\caption{BRIGHT nDCG@10 by domain and model tier (4B, 8B, 9B, 14B), the full breakdown behind Table~\ref{tab:bright-main}. \baselinename{} vs.\ \methodname{}.}
\label{fig:app-bright-domains}
\end{figure}

\begin{figure}[ht]
\centering
\includegraphics[width=0.75\textwidth]{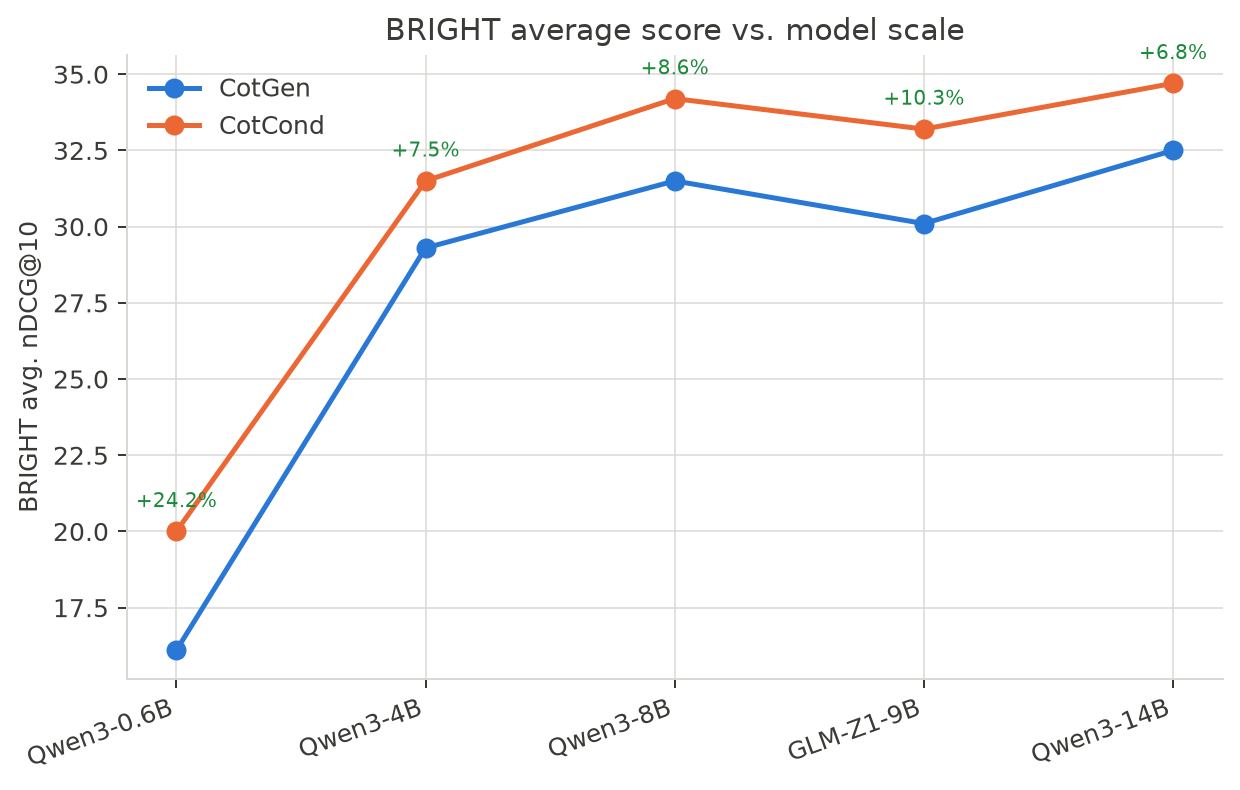}
\caption{BRIGHT average nDCG@10 vs.\ model scale (4B, 8B, 9B, 14B). \methodname{} is above \baselinename{} at every size.}
\label{fig:app-bright-scale}
\end{figure}

Figure~\ref{fig:app-bright-domains} shows the BRIGHT comparison by individual domain and model tier. \methodname{} leads on the 12-domain average at every tier, though not on every domain within a tier; the regression domains and counts are listed in Section~\ref{sec:results-passage}. Figure~\ref{fig:app-bright-scale} plots the same averages against model size.

\begin{figure}[ht]
\centering
\includegraphics[width=0.75\textwidth]{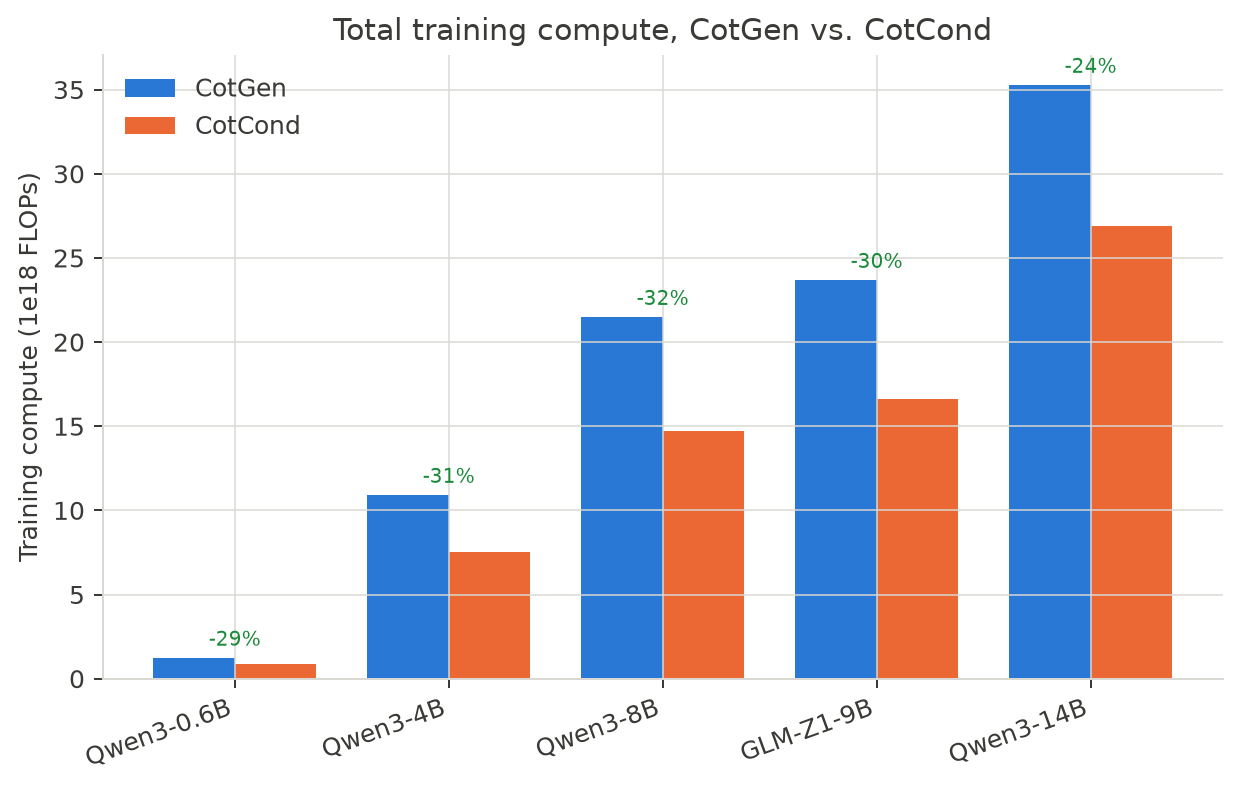}
\caption{Reported training-compute estimates by tier (4B, 8B, 9B, 14B). The paired records show lower estimated compute and higher average BRIGHT scores under \methodname{}. These estimates use the $6ND$ proxy discussed in Section~\ref{sec:efficiency}.}
\label{fig:app-flops-tier}
\end{figure}

\subsection{Table reranking: in-domain comparison}
\label{sec:appendix-indomain}

The in-domain comparison is given in the main text (Table~\ref{tab:table-reranking}). On the 0--100 nDCG@10 scale used throughout, $\bar\Delta_{\mathrm{OOD}} = 3.05$ and $\bar\Delta_{\mathrm{ID}} = 1.02$, giving $\bar\Delta_{\mathrm{OOD}} - \bar\Delta_{\mathrm{ID}} = 2.03$ (Section~\ref{sec:results-table}).

\subsection{Generation length}
\label{sec:appendix-generation-length}

On the analyzed Qwen3-8B passage-reranking sample (899 matched query-passage pairs across all twelve BRIGHT domains), \methodname{} generates a mean of 198 words per trace against \baselinename{}'s 312 words, a baseline-to-\methodname{} word-count ratio of $1.58$. The reduction holds in all twelve BRIGHT domains, with the largest ratio on AoPS ($2.3\times$) and the smallest on Biology ($1.4\times$).

\subsection{Inference efficiency}
\label{sec:appendix-inference}

\Methodname{} has a lower recorded inference time for each model pair in the reported BRIGHT workload. The records do not specify timed workload counts or repeated timing trials. Elapsed times are reported in Table~\ref{tab:consolidated-results-full} and summarized in Section~\ref{sec:efficiency}.

\subsection{Training loss curves}
\label{sec:appendix-loss-curves}

\begin{figure}[ht]
\centering
\includegraphics[width=0.6\textwidth]{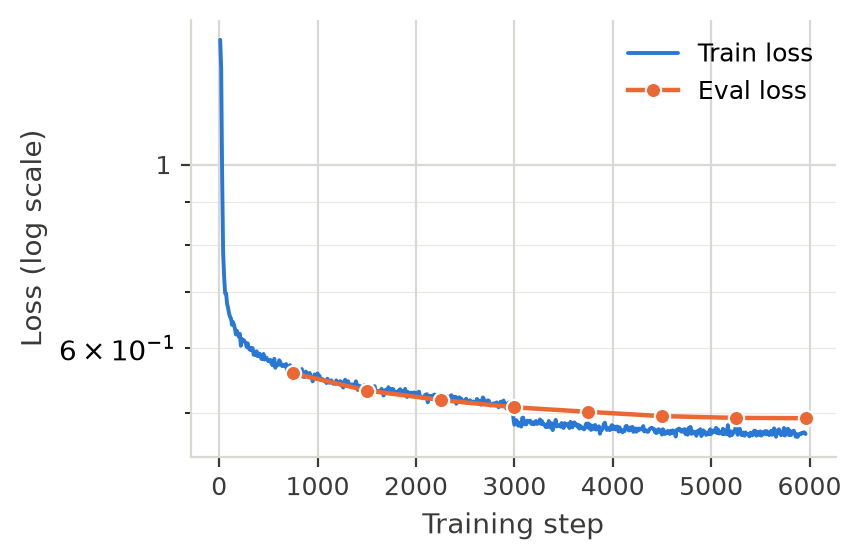}
\caption{Training/eval loss -- Qwen3-4B, \baselinename{}.}
\label{fig:loss-qwen34b-cotgen}
\end{figure}

\begin{figure}[ht]
\centering
\includegraphics[width=0.6\textwidth]{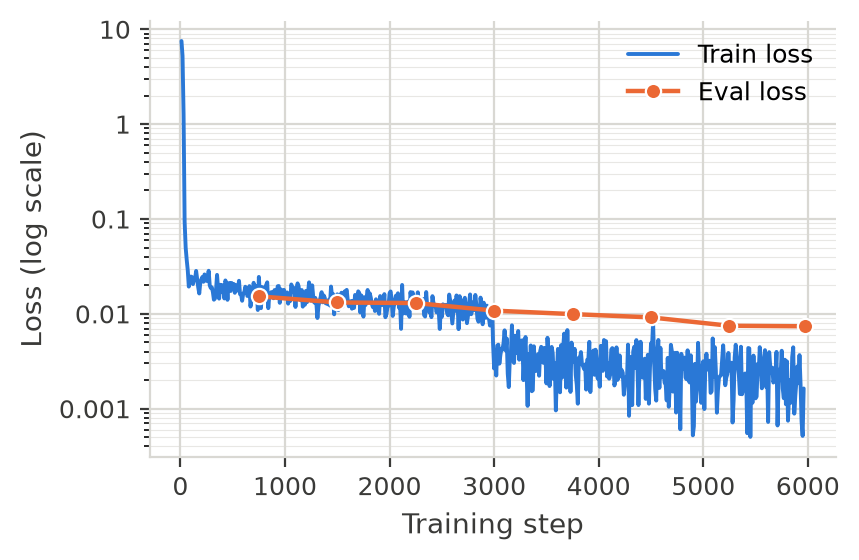}
\caption{Training/eval loss -- Qwen3-4B, \methodname{}.}
\label{fig:loss-qwen34b-cotcond}
\end{figure}

\begin{figure}[ht]
\centering
\includegraphics[width=0.6\textwidth]{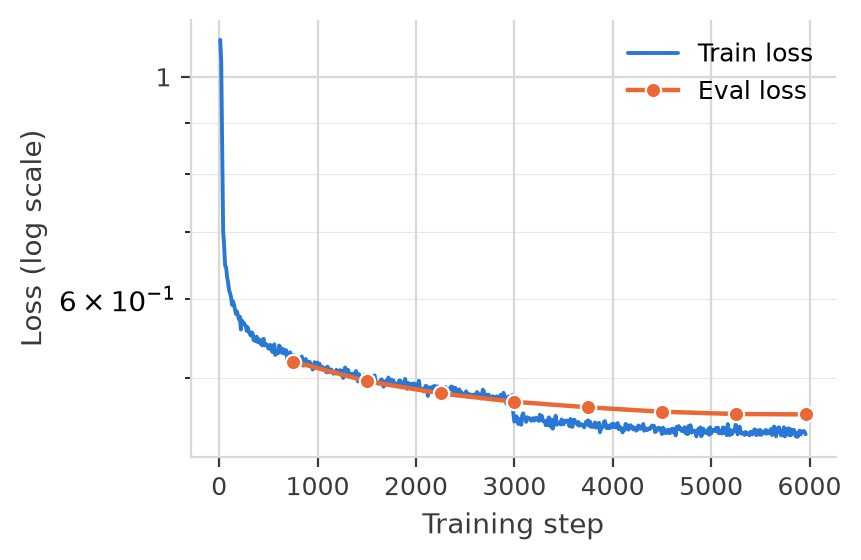}
\caption{Training/eval loss -- Qwen3-8B, \baselinename{}.}
\label{fig:loss-qwen38b-cotgen}
\end{figure}

\begin{figure}[ht]
\centering
\includegraphics[width=0.6\textwidth]{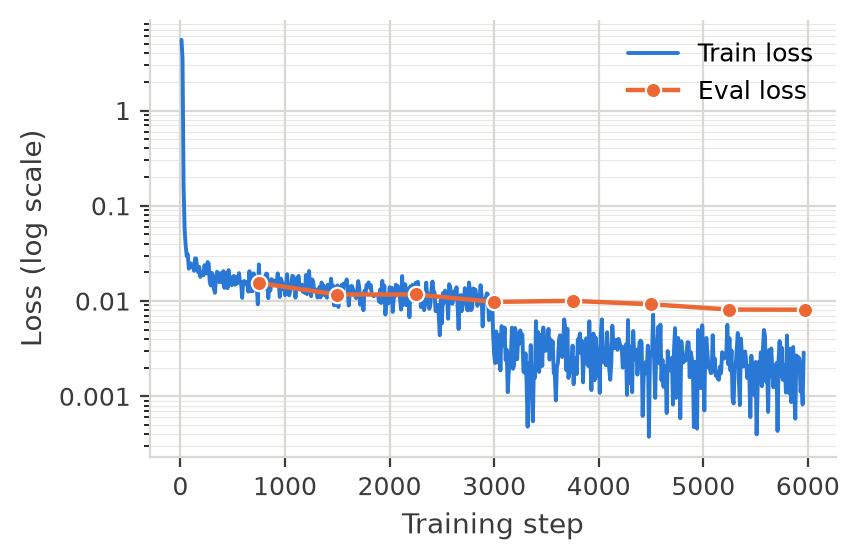}
\caption{Training/eval loss -- Qwen3-8B, \methodname{}.}
\label{fig:loss-qwen38b-cotcond}
\end{figure}

\begin{figure}[ht]
\centering
\includegraphics[width=0.6\textwidth]{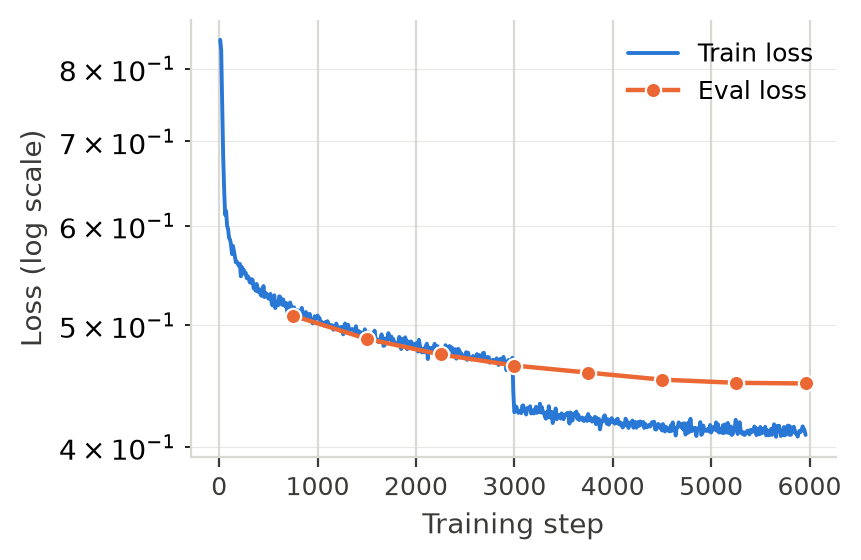}
\caption{Training/eval loss -- GLM-Z1-9B, \baselinename{}.}
\label{fig:loss-glmz19b-cotgen}
\end{figure}

\begin{figure}[ht]
\centering
\includegraphics[width=0.6\textwidth]{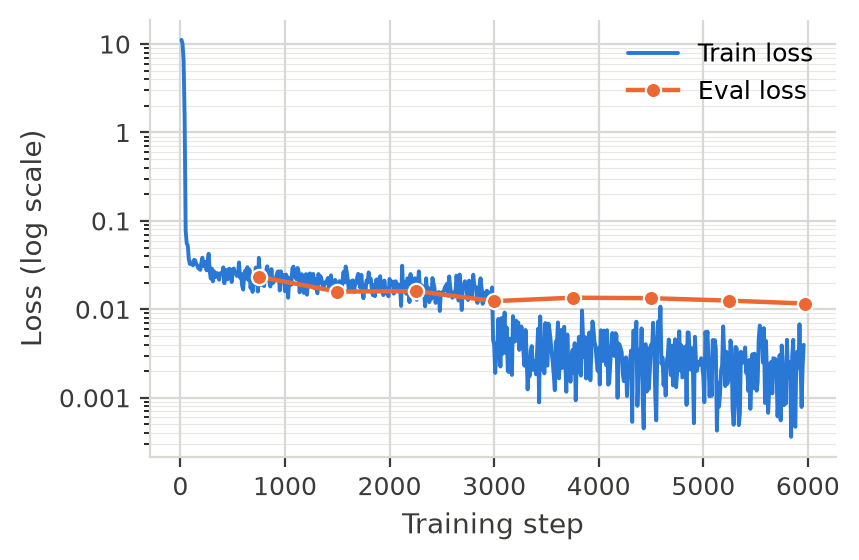}
\caption{Training/eval loss -- GLM-Z1-9B, \methodname{}.}
\label{fig:loss-glmz19b-cotcond}
\end{figure}

Figures~\ref{fig:loss-qwen34b-cotgen}--\ref{fig:loss-glmz19b-cotcond} give training and evaluation loss curves for the Qwen3-4B, Qwen3-8B, and GLM-Z1-9B conditions; we do not have corresponding curves for Qwen3-14B. In interpreting these curves, absolute loss values across methods are not directly comparable because the supervised token sets differ (Section~\ref{sec:selective-supervision}). \baselinename{} computes loss over both reasoning and answer tokens; \methodname{} computes loss over answer tokens only. Differences in supervised token composition and weighting prevent the absolute loss levels from serving as a direct comparison of convergence quality across methods.

At the epoch boundary (step $\approx$3{,}000 of 6{,}000), \methodname{}'s training loss drops roughly an order of magnitude while its evaluation loss barely moves (Figure~\ref{fig:loss-qwen38b-cotcond}), a pattern consistent with the model memorizing answers for its specific training examples rather than improving its general answer-prediction ability. \Baselinename{}'s training loss shows a comparable but visibly smaller drop at the same boundary (Figure~\ref{fig:loss-qwen38b-cotgen}). This is consistent with the supervised answer being recoverable from the training context under \methodname{} (Section~\ref{sec:reasoning-transformation}), and it means the near-zero training loss should not be read as evidence that the model has learned a general answer-prediction skill from the second epoch onward.

\subsection{Generated Output Length}
\label{sec:appendix-generated_output}
\begin{table}[t]
    \centering
    \small
    \caption{\textbf{Generated output length per query} (tokens), table reranking. Both methods generate reasoning before answering. \methodname{} consistently produces shorter outputs.}
    \label{tab:output-length}
    \begin{tabular}{lccc}
    \toprule
    \textbf{Benchmark} & \textbf{\baselinename{}} & \textbf{\methodname{}} & \textbf{Ratio} \\
    \midrule
    SQA       & 3{,}318.5 & 2{,}145.3 & $1.55\times$ \\
    TAT-QA    & 3{,}704.2 & 2{,}473.4 & $1.50\times$ \\
    TabFact   & 3{,}716.1 & 2{,}454.3 & $1.51\times$ \\
    NQ-Tables & 2{,}403.9 & 1{,}361.7 & $1.77\times$ \\
    \bottomrule
    \end{tabular}
    \vspace{-1.0em}
\end{table}
Table~\ref{tab:output-length} reports the average number of generated tokens per query for table reranking. Across all four benchmarks, \methodname{} produces substantially shorter outputs than \baselinename{}, with reductions ranging from $1.50\times$ to $1.77\times$. The largest reduction occurs on NQ-Tables, where the average output length decreases from 2,403.9 to 1,361.7 tokens. These results show that the shorter-generation pattern observed in passage reranking also extends to table reranking.

On passage reranking, the same $1.58\times$ average reduction (Section~\ref{sec:efficiency}) holds across all twelve BRIGHT domains, with the largest ratio on AoPS ($2.3\times$) and the smallest on Biology ($1.4\times$).

\section{Additional Analysis}
\label{sec:appendix-analysis}

The surface-style and voice-proximity analyses are given below (Appendix~\ref{sec:analysis-style}).

\subsection{Exploratory generative reasoning}
\label{sec:appendix-generative}

The available exploratory summary covers five model families and seven mathematical-reasoning benchmarks, with 29 reported model--benchmark comparisons. \Methodname{} generates fewer reasoning tokens in 25 of 29 comparisons, with median baseline-to-\methodname{} length ratio $1.5$. It has lower pass@1 in 23 of 29 comparisons, with mean accuracy difference $-5.9$ percentage points.

This summary shows that shorter outputs do not imply better accuracy outside the reranking tasks. The supplied source does not include the 29 individual score cells, model and benchmark identities, or the complete training and evaluation configurations. We therefore retain these as an aggregate exploratory finding and do not treat them as a fully specified cross-domain experiment.

\subsection{Surface Style and Voice}
\label{sec:analysis-style}

One might ask whether \methodname{}'s improvement arises from more varied or creative phrasing. On the contrary, \methodname{} is \emph{more} formulaic. On the 899 matched passage pairs, \methodname{} reuses the same six-word opening in 94\% of generations versus 45\% for \baselinename{}, and its within-domain trigram Jaccard similarity is also higher (0.049 vs.\ 0.040). The improvements therefore cannot be attributed to greater surface diversity; \methodname{}'s gains come despite more stereotyped phrasing.

On table reranking, we measure lexical proximity to the base model using length-matched Jensen--Shannon divergence over 1{,}370 matched questions. \Methodname{} is 8\% closer to the base model's token distribution than \baselinename{} (JSD 0.232 vs.\ 0.252, Wilcoxon $p=1.9\times10^{-22}$), with higher bigram overlap (0.298 vs.\ 0.263). This suggests that removing trace imitation allows the student to retain more of its pretrained voice rather than adopting the teacher's stylistic patterns. \Naivename{} shows a much larger divergence from base (JSD 0.373), indicating that the proximity is specific to \methodname{}'s training with teacher context, not a general property of answer-only objectives.


\end{document}